\documentclass{article}
\usepackage{ijcai23}

\usepackage{times}
\usepackage{soul}
\usepackage{url}
\usepackage[hidelinks]{hyperref}
\usepackage[utf8]{inputenc}
\usepackage[small]{caption}
\usepackage{subcaption}

\usepackage{graphicx}

\usepackage{amsthm}

\usepackage{booktabs}
\usepackage{algorithm}
\usepackage{algorithmic}
\usepackage[switch]{lineno}
\usepackage{tikz}
\usetikzlibrary{shapes, arrows}
\usepackage{verbatim}

\usepackage{todonotes}

\usepackage{amsmath} 
\usepackage{amssymb}  

\title{Risk-sensitive Actor-free Policy via Convex Optimization}

\author{
Ruoqi Zhang$^1$
\and
Jens Sjölund$^1$\and
\affiliations
$^1$Uppsala University\\
\emails
firstname.lastname@it.uu.se
}

\begin{document}

\maketitle

\begin{abstract}
Traditional reinforcement learning methods optimize agents without considering safety, potentially resulting in unintended consequences.  In this paper, we propose an \emph{optimal} actor-free policy that optimizes a risk-sensitive criterion based on the conditional value at risk. The risk-sensitive objective function is modeled using an input-convex neural network ensuring convexity with respect to the actions and enabling the identification of globally optimal actions through simple gradient-following methods. 
Experimental results demonstrate the efficacy of our approach in maintaining effective risk control.
\end{abstract}

\section{Introduction}
Over the past decade, reinforcement learning (RL) has achieved notable advancements \cite{mnih2015human}.
Nonetheless, traditional RL agents interact with their environment without accounting for safety, potentially leading to unintended and severe consequences in real-world applications.
Safe RL addresses these concerns by ensuring that the learning process is both effective and safe. An intuitive way is to learn the policy subject to safety constraints, accounting for both parametric and inherent uncertainties within the model and its environment \cite{garcia2015comprehensive_safe_survey}.

In this paper, we explore this approach by 
training a policy that optimizes a risk-sensitive criterion based on the conditional value at risk (CVaR). 
Traditional RL algorithms typically aim to maximize the expected value over all future cost-returns \cite{ha2021learning}. 
By emphasizing the tail of the future cost-return distribution, 
our learned policy explicitly penalizes infrequent occurrences of catastrophic events.

Additionally, we propose an actor-free architecture in which the action is implicitly defined as the solution to a convex optimization problem approximating the risk-sensitive criterion.  
This eliminates the need for incremental actor learning, which often necessitates hyperparameter tuning and tricks to stabilize the training process. With our actor-free approach, the policy aligns optimally with the approximated criterion. 
This deviates from prior research on CVaR-based safe RL \cite{yang2021wcsac,tang2020worst}, which uses a neural network to approximate the actor.  
Key to our approach is to parameterize the risk-sensitive objective function using an input-convex neural network \cite{amos2017input}, ensuring convexity with respect to the actions (inputs). Consequently, simple gradient-following techniques can be used to find a globally optimal action.

\section{Risk-sensitive Actor-free Policy}
In this study, we focus on learning a risk-sensitive actor-free policy within a safety-constrained framework. 
The aim of the agent is to optimize future returns, maintaining compliance with safety cost constraints. 
The risk-sensitive objective function is structured using an input-convex neural network \cite{amos2017input}, guaranteeing convexity with respect to the actions (inputs). As a result, a globally optimal action can be identified.

\subsection{Constrained Markov Decision Processes}
\label{sec:cmdp}
We model the RL agent and its environment as a constrained Markov Decision Process (CMDP), represented by 
a tuple $(\mathcal{S}, \mathcal{A},\mathcal{P}, r, c, d, \gamma)$ where $\mathcal{S} \in \mathbb{R}^{d_\mathcal{S}}$ is the state space, $\mathcal{A} \in \mathbb{R}^{d_\mathcal{A}}$ is the action space,  $\mathcal{P}: \mathcal{S} \times \mathcal{A} \times \mathcal{S} \rightarrow \mathbb{R}$ is the probabilistic transition function, $r:  \mathcal{S} \times \mathcal{A} \rightarrow \mathbb{R}$ is the immediate reward function, $c:  \mathcal{S} \times \mathcal{A} \rightarrow \mathbb{R}$ is the immediate cost function, $d \in \mathbb{R}$ is the safety threshold 
and $\gamma \in (0.0,1.0)$ is the discount factor.\looseness=-1
The goal of the agent under the CMDP framework is to learn a policy that maximizes the expected return given an upper bound $d$ on the  (safety violation) cost,
\small
\begin{equation}
    \begin{aligned}
        & \underset{\pi}{\text{maximize}}
        & & \mathbb{E}_{s_t \sim \rho_\pi, a_t \sim \pi(s_t)} \left[ 
    \sum_t \gamma^t r(s_t, a_t)
    \right ] \\
        & \text{subject to}
        & &\mathbb{E}_{s_t \sim \rho_\pi, a_t \sim \pi(s_t)}    \left[ 
    \sum_t \gamma^t c(s_t, a_t)
    \right ]  \leq d.
    \end{aligned}
    \label{eq:criterion}
\end{equation}
\normalsize
where $\rho_\pi$ is the stationary distribution over the state space $\mathcal{S}$ under the policy $\pi$.

\subsection{Safety Critic with Conditional Value at Risk}
\label{subsec:cvar}
In CMDP, the safety violation costs are usually the (in)finite-horizon discounted future cost-return as shown in \eqref{eq:criterion}.
However, only considering the expected value is insensitive to potentially hazardous events: policy gradient methods prefer a policy with lower cost, but also higher variance, over a policy with slightly higher cost but much lower variance. In other words, since higher variance amounts to higher risk, the policy is not risk-averse.
To incorporate risk, we replace the expectation in the safety violation cost with the Conditional Value at Risk (CVaR) \cite{artzner1999cvar}, a widely recognized risk measure that quantifies the amount of tail risk. More precisely, CVaR$_\alpha$ is defined as the expected reward of the worst $\alpha$-percentile cases,
\small
\begin{align}
    \Gamma_c^\pi(s,a,\alpha) \doteq \text{CVaR}_\alpha^\pi(C) = \mathbb{E}_{p_\pi}
    \left [
    C \mid C \geq F_C^{-1}(1-\alpha)
    \right ]
\end{align}
\normalsize
where $\alpha\in (0,1]$ is used to define the risk level, $C$ is a random variable and $F_C^{-1}(1-\alpha)$ is the $\alpha$-percentile. 
Calculating the CVaR measure directly for lengthy time horizons using, for example, sampling would be excessively costly \cite{tamar2015optimizing}.  
Instead, we follow \cite{tang2020worst} and model the distribution of cost-return $C(s,a)$ as a Gaussian distribution $\mathcal{N}\left(Q^\pi_c(s,a), \sigma^2_c (s,a)\right)$ where $Q_c^\pi (s,a)=\mathbb{E}_\pi[\sum_{i=t}^\infty\gamma^{i-t}c(s_i,a_i)\,\vert\, s_t=s, a_t=a
]$ is the expected future cost-return and $\sigma^2_c(s,a)$ its variance. This Gaussian distribution leads to a closed-form CVaR measure of future cost-return \cite{yang2021wcsac,tang2020worst}
\begin{align}
    \Gamma_c^\pi(s,a,\alpha)  = Q_c^\pi(s,a) + \frac{ \phi(\alpha)}{ \Phi(\alpha)} \sigma_c^\pi(s,a),
    \label{eq:cvar}
\end{align} where $\phi$ is the standard normal distribution, and $\Phi(\cdot)$ is its CDF.
Our risk-sensitive criterion can be written as,
\begin{equation}
    \begin{aligned}
        & \underset{\pi}{\text{maximize}}
        & & Q_r^\pi(s,a)\\
        & \text{subject to}
        & & Q_c^\pi(s,a) + \frac{ \phi(\alpha)}{ \Phi(\alpha)} \sigma_c^\pi(s,a) \leq d,
    \end{aligned}
    \label{eq:cvar-criterion}
\end{equation} where $Q_r^\pi(s,a) = \mathbb{E}_\pi[\sum_{i=t}^\infty\gamma^{i-t}r(s_i,a_i)\vert s_t=s, a_t=a]$ is the expected future return.
To learn the mean and variance of cost-to-go, a distributional critic is learned with 2-Wasserstein distance as the loss function \cite{tang2020worst}.

\subsection{\emph{Optimal} Actor-free Policy via Input Convex Neural Network}
\label{sec:actor-free}
Policy gradient algorithms typically feature an actor-critic structure, utilizing two distinct neural networks known as the actor and the critic \cite{fujimoto2018TD3}. The critic estimates the reward-to-go or cost-to-go and the actor seeks to infer the action $a$ to maximize the estimation from the critic.
However, if the \emph{optimal} action w.r.t the critic can be easily identified, the need for modeling the actor is eliminated. This can be achieved through parameterization of the reward with Partially Input Convex Neural Networks (PICNNs) \cite{amos2017input}. We utilize two PICNNs, one to approximate $-Q_r(s,a)$ and another to estimate the cost-return distribution $Q_c(s,a)$ and $\sigma_c(s,a)$. 
In this way, 
the policy is \emph{actor-free} since the optimal action $a^*$ can be determined directly by minimizing 
\small
\begin{align}
 a^* = \arg\min_a -Q_r(s,a) + \kappa \max
    \{0, \Gamma_c(s_t,a_t)-d\}\label{eq:augmented_lagrangian}
\end{align}
\normalsize
where $\kappa\geq 0$ is a hyperparameter. This optimization problem is a nonsmooth exact penalty formulation of the constrained problem in \eqref{eq:cvar-criterion}. It is well-known that for sufficiently large $\kappa$ the two problems have the same solution \cite{wright1999numericalopt}. Moreover, since this problem is convex, finding a globally optimal action is tractable. 

\begin{figure}[ht]
   \centering
   \includegraphics[width=0.6\linewidth]{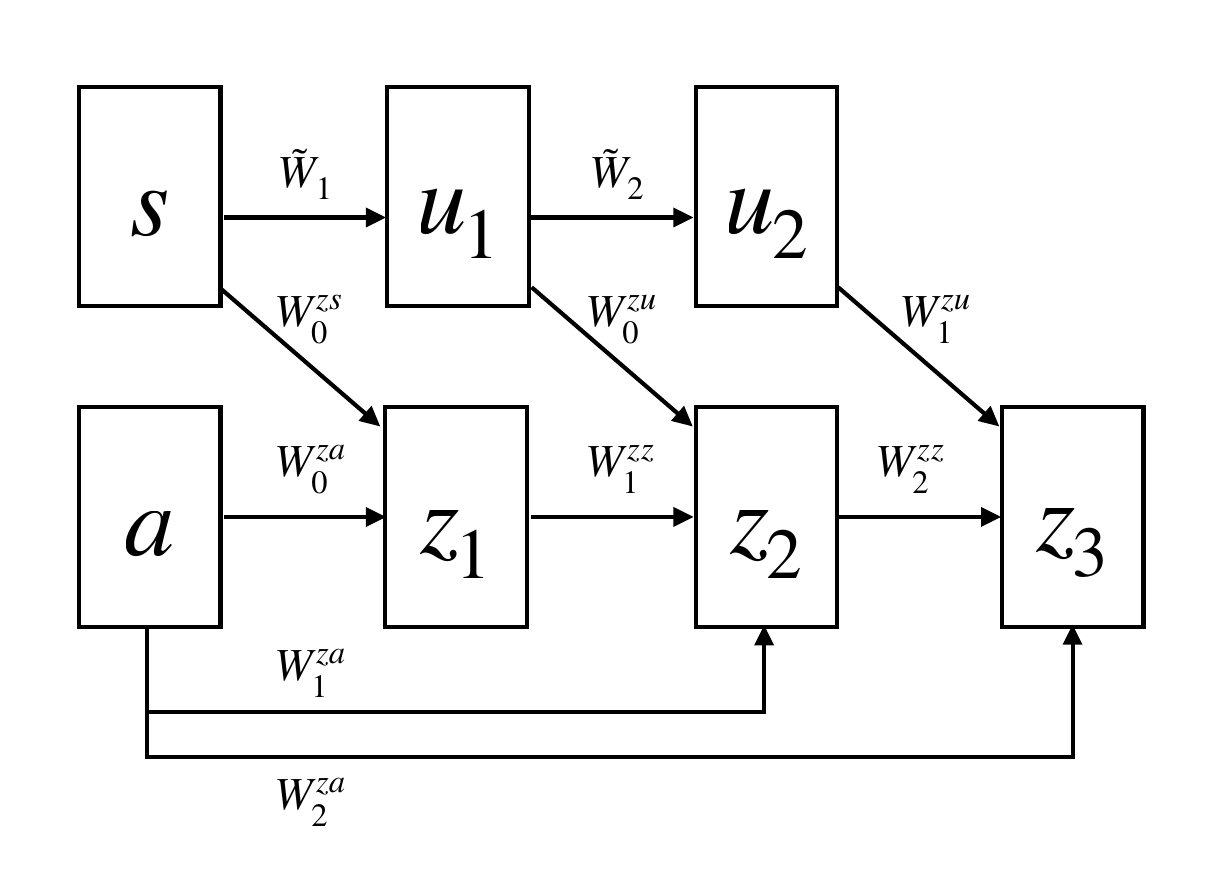}
   \caption{The partially input convex neural network (PICNN) structure used in the paper.}
   \label{fig:convex_net_strucure}
\end{figure}

We first define the PICNN over state-action pairs $f(s,a;\theta)$ where $f$ is convex in action $a$ but not convex in state $s$. Figure~\ref{fig:convex_net_strucure} illustrates the simple convex network structure used in our paper. As shown in the figure, output $z_3$ can be calculated by forwarding the network, 
\begin{align}
    u_0 &= s, ~~u_{i+1} = \tilde{g}_i(\tilde{W_i} u_i+\tilde{b}_i), &i=0, 1  \nonumber  \\
    z_1 &= g_0(W_0^{zs}s + W_0^{za}a + b_0), \nonumber \\
    z_{i+1} &= g_i(W_i^{zu}u_i + W_i^{za}a + W_i^{zz}z) , &i=1, 2
\end{align}
where $W$ are weight matrices, $b$ are bias terms, $g$ is the nonlinear activation function and $z_3$ is the output of the network which is made convex in the input $a$ by restricting the weight matrices $W_0^{za}$ and $W_i^{zz}$ to be non-negative and the activation function $g$ to be convex and non-decreasing, e.g. a rectified linear unit (ReLU). 

\section{Experiment}
\begin{figure}[ht]
   \centering
   \includegraphics[width=0.55\linewidth]{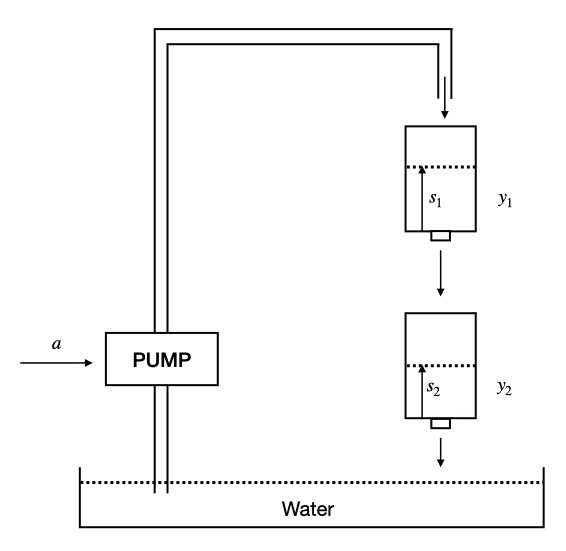}
   \caption{The water tank level control task.}
   \label{fig:watertank-process}
\end{figure}

\begin{figure*}
    \centering
     \begin{subfigure}[b]{0.25\textwidth}
         \centering
         \includegraphics[width=\textwidth]{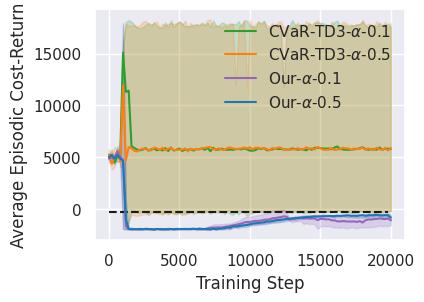}
         \caption{Training cost-return}
         \label{fig:training-cost}
     \end{subfigure}
     \begin{subfigure}[b]{0.25\textwidth}
         \centering
         \includegraphics[width=\textwidth]{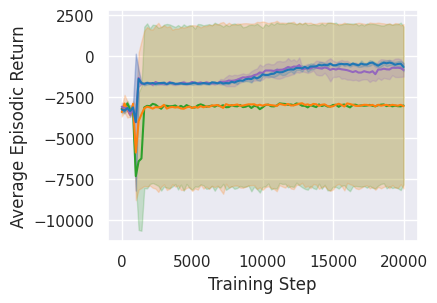}
         \caption{Training return}
         \label{fig:training-return}
     \end{subfigure}
     \begin{subfigure}[b]{0.45\textwidth}
         \centering
         \includegraphics[width=\textwidth]{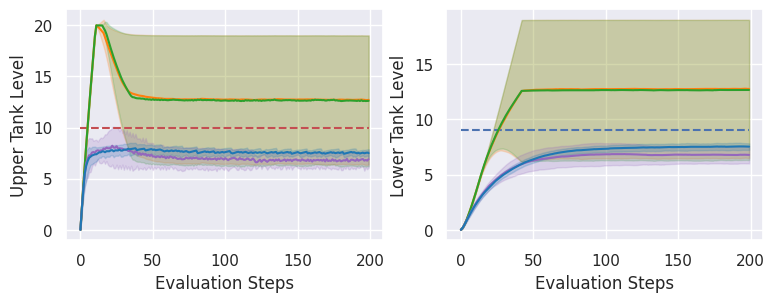}
         \caption{Evaluation}
         \label{fig:eva}
     \end{subfigure}
    \caption{Performance comparison between our method with CVaR-TD3 with different risk values $\alpha=0.1, 0.5$ during training with mean (solid lines) and  $\pm 1$ standard deviation (shaded area) of 5 runs. (The cost-return threshold $d$ is in \emph{dashed black}, the ``critical'' level is in \emph{dashed red},  and the goal level is in \emph{dashed blue}.)}
    \label{fig:training}
\end{figure*}
The above method was evaluated in simulation on a continuous control task: cascade water tank level control. 
As depicted in Figure~\ref{fig:watertank-process}, the task is to maintain a specific water level in the lower tank by changing the input signal $a_t$ represents the voltage to the pump.
The state of the system includes the heights of the two tanks, $s(t) = \begin{bmatrix}s_1(t) & s_2(t)\end{bmatrix}^T$ and the output is $y(t) = s_2(t)$.
The reward function contains two parts, one related to the distance between the reference signal and the output, and the other to the cost incurred by a critical event. To introduce a risk element, we define a ``critical'' event as the level of the upper tank exceeding $l_\text{crit}=10$ cm. Thus the reward and cost functions are defined as $r(s,a) = -\vert s_2-g\vert$, and $c(s,a)=s_1-l_\text{crit}$, respectively. The long-term safety threshold $d$ is set to $-250$. The system was discretized using the Euler method in the simulation, with a sampling period of 2 seconds.  Unlike a real-world tank, there are no upper bounds for $s_1, s_2$ in the simulation.

Our learning algorithm to update reward critic and safety critic is based on Twin Delayed Deep Deterministic policy gradient algorithm (TD3) \cite{fujimoto2018TD3} to avoid overestimating Q-values.
As mentioned in Section~\ref{subsec:cvar}, we use a reward critic and a distributional safety critic. Further, the policy is \emph{actor-free}; the optimal action is found by a gradient descent algorithm, Adam \cite{kingma2015adam}.

We compare our method with its actor-critic version, CVaR-TD3, using a standard neural network instead of the PICNNs for actor, critic, and safety critic.  
The training results are illustrated in Figure~\ref{fig:training} and Figure~\ref{fig:training-cost} with two metrics, average episodic returns, and average episodic cost-returns.  It can be observed that CVaR-TD3 has a much higher variance, potentially attributed to the neural network-structured actor getting stuck in poor local minima, while our approach can identify the globally optimal action, aided by the PICNNs. 
Careful tuning of the hyperparameters could potentially mitigate this. 
Figure~\ref{fig:eva} shows the evaluation of the learned policy. To highlight the differences between the methods, the state $s_1, s_2$ is clipped with a maximum value $20$.
We observe that our method with $\alpha = 0.1$ exhibits a more conservative behavior, striving to maintain a safe distance from the critical level (\emph{dashed red}). Consequently, our approach is also slightly farther from the goal level (\emph{dashed blue}), which is close to the critical level.

\section{Conclusions}
We proposed a risk-sensitive actor-free policy with a CVaR criterion. The criterion is parameterized with input-convex neural networks ensuring convexity with respect to the actions. Thus, the globally optimal action can be found easily by simple gradient-descent methods. In the paper, future return and cost-return are approximated by a Gaussian distribution in order to get a closed-from of CVaR. Future research could explore a more general distribution.




\bibliographystyle{named}
\bibliography{cvxbib}

\end{document}